\documentclass[11pt,a4paper]{article}
\usepackage[hyperref]{ranlp2023}
\usepackage{times}
\usepackage{latexsym}
\usepackage{graphicx}

\usepackage{microtype}
\usepackage{booktabs}  
\usepackage{multirow}  

\aclfinalcopy 

\title{Multilingual Hope Speech Detection: A Comparative Study of Logistic Regression, mBERT, and XLM-RoBERTa with Active Learning}

\author{
  \textbf{Abiola T. O.\textsuperscript{1}},
  \textbf{Abiodun K. D.\textsuperscript{2}},
  \textbf{Olumide O. E.\textsuperscript{1}},
  \textbf{Adebanji O. O.\textsuperscript{1}},\\
  \textbf{Hiram Calvo O.\textsuperscript{1}},
  \textbf{Sidorov Grigori.\textsuperscript{1}}
\\
  \textsuperscript{1}Instituto Politécnico Nacional, Centro de Investigación en Computación, CDMX, Mexico.\\
  \textsuperscript{2}Ekiti State University, Ado-Ekiti, Nigeria.
\\
\small{\href{mailto:sidorov@cic.ipn.mx}{sidorov@cic.ipn.mx}
 }
}

\date{}

\begin{document}
\maketitle
\begin{abstract}
Hope speech language that fosters encouragement and optimism plays a vital role in promoting positive discourse online. However, its detection remains challenging, especially in multilingual and low-resource settings. This paper presents a multilingual framework for hope speech detection using an active learning approach and transformer-based models, including mBERT and XLM-RoBERTa. Experiments were conducted on datasets in English, Spanish, German, and Urdu, including benchmark test sets from recent shared tasks. Our results show that transformer models significantly outperform traditional baselines, with XLM-RoBERTa achieving the highest overall accuracy. Furthermore, our active learning strategy maintained strong performance even with small annotated datasets. This study highlights the effectiveness of combining multilingual transformers with data-efficient training strategies for hope speech detection. 
\end{abstract}

\section{Introduction}
The rapid proliferation of social media has revolutionised the way people communicate, enabling the real-time exchange of thoughts, experiences, and emotions. While these platforms have become essential for digital interaction, they are often dominated by harmful content, including hate speech, cyberbullying, and misinformation. In response, there has been growing academic and societal interest in identifying and amplifying positive discourse, particularly "hope speech," which encompasses messages of encouragement, optimism, reassurance, and motivation.

Hope speech plays a crucial role in fostering resilience, reducing negative emotions such as anxiety and loneliness, and promoting mental well-being within online communities. However, the automatic detection of such speech remains a challenging task, especially given the linguistic and cultural diversity of social media users. Recent studies have shown promising results in detecting hope speech across various languages, including Spanish, English, Tamil, and Malayalam \cite{ahmad2024posivox, eyob2024enhancing, armenta2024ometeotl}. These works have employed a variety of machine learning techniques, from traditional models like Support Vector Machines (SVM) and Random Forest (RF) to state-of-the-art transformer models such as BERT, which have proven highly effective in capturing the semantic nuances of positive discourse. As attention has expanded to low-resource and multilingual settings, models like BERT, mBERT, and XLM-RoBERTa have shown promise in capturing cross-lingual semantic nuances, significantly outperforming traditional approaches \cite{ahmad2024posivox, armenta2024ometeotl}.

Recent shared tasks and studies have explored hope speech detection in languages such as Spanish, Tamil, Urdu, and Malayalam, but substantial gaps remain, particularly regarding scalable, annotation-efficient methods suitable for underrepresented languages. This study addresses these limitations by introducing a multilingual framework for hope speech detection that incorporates an active learning paradigm. By iteratively selecting the most informative samples for learning, our method reduces the burden on small annotated data while maintaining high classification performance.

This research evaluates the effectiveness of three models, Logistic Regression, mBERT, and XLM-RoBERTa, on four languages: English, Spanish, German, and Urdu. Our experiments utilise both development and test datasets, including those from the HOPE shared tasks, and benchmark against existing literature. The study contributes to the growing body of work on computational hope speech detection by demonstrating the effectiveness of transformer models in multilingual and low-resource settings, validating the efficiency of active learning in enhancing model performance with a small dataset.

\section{Literature Review}
Text detection and classification have seen growing interest in NLP, with researchers exploring various traditional \citealp{ojo2021performance, Ojo2020, Sidorov2013} and deep learning models \citealp{Aroyehun2018, Ashraf2020, Han2021, hoang2022combination, Poria2015, muhammad2025brighterbridginggaphumanannotated}, as well as improved classifiers \citealp{Abiola2025a, Kolesnikova2020, ojo2024doctor, adebanji2022, Abiola2025b}. Recently, hope speech detection has emerged as a focused area, with multilingual datasets and transformer-based approaches gaining attention \citealp{sidorov2024mindhope, balouchzahi2025urduhope, balouchzahi2025polyhopem, balouchzahi2023polyhope, sidorov2023regrethope, garcia2024overviewiberlef, garcia2023lgthope}.

Recent studies such as \cite{arif2024analyzing} and \cite{sharma2025ensemble} have explored hope speech detection through diverse methodologies. \cite{arif2024analyzing} applied NLP tools like LIWC, NRC-emotion-lexicon, and VADER to extract psycholinguistic and emotional features, achieving strong classification performance using tuned LightGBM and CatBoost models. Meanwhile, \cite{sharma2025ensemble} proposed an ensemble model combining LSTM, mBERT, and XLM-RoBERTa for multilingual hope speech detection in English, Kannada, Malayalam, and Tamil, reporting superior results with weighted F1-scores of 0.93, 0.74, 0.82, and 0.60, respectively.

\cite{divakaran2024hope} defined hope speech as content promoting positivity and community support, participating in the IberLEF 2024 shared task using ML and TL approaches like TF-IDF features and fine-tuned BERT models. Their systems achieved macro F1 scores up to 0.82 in binary classification and 0.64 in multiclass settings. Similarly, \cite{ahmad2025multilingual} focused on English and Urdu, proposing a translation-based, multilingual BERT approach. Their models achieved accuracies of 87\% for English and 79\% for Urdu, outperforming traditional baselines and highlighting BERT’s effectiveness for low-resource language tasks.

\cite{arunadevi2024ml} explored hope speech detection in Spanish as part of the HOPE 2024 shared task, employing logistic regression with TF-IDF and count vectorizer features. Their model achieved a macro-F1 score of 0.4161, placing 15th, demonstrating that even traditional ML approaches have potential in low-resource and multilingual settings. Meanwhile, \cite{nath2025bonghope} addressed the underrepresentation of Bengali in hope speech research by introducing BongHope, the first annotated Bengali dataset for binary classification of hope speech. Their work not only fills a critical resource gap but also enables further computational analysis of positive discourse in Bengali social media.

\cite{armenta2024ometeotl} proposed custom BERT models tailored for multilingual sentiment tasks using the HopeEDI (Spanish) and PolyHope (Spanish \& English) datasets in the HOPE at IberLEF 2024 shared task. Their models achieved strong results, ranking fourth, sixth, and eighth across various subtasks, demonstrating the value of language\-specific pretraining. Similarly, \cite{eyob2024enhancing} highlighted the effectiveness of BERT in detecting hope speech on Twitter, outperforming traditional ML models like SVM and Random Forest with a macro-F1 score of 0.85. Both studies affirm the strength of transformer models in multilingual and cross-cultural hope speech detection.

\section{Methodology}
To address the challenges of detecting hope speech across diverse and low-resource languages, we propose a robust methodology that leverages models within an active learning framework. Our approach involves iterative data addition guided by model uncertainty, enabling efficient use of limited labelled data. We employ a Logistic Regression model and multilingual pre-trained language models such as mBERT and XLM-RoBERTa, fine-tuned on curated datasets including English, Urdu, Spanish and German, to enhance cross-lingual generalizability. The following subsections detail our dataset preparation, model architecture, active learning, and evaluation protocols.

\subsection{Datasets}
We utilised multilingual datasets across four languages: English, German, Spanish, and Urdu as given by the PolyHope-M task organisers \cite{balouchzahi2025polyhopem, balouchzahi2023polyhope}. Each dataset was annotated for binary classification with two labels: \textit{Hope} and \textit{Not Hope}. The distribution of these labels in both training and development sets is shown in Table~\ref{tab:dataset-distribution}.

\begin{table}[h!]
\centering
\begin{tabular}{llccc}
\toprule
\textbf{Language} & \textbf{Split} & \textbf{Hope} & \textbf{Not Hope} & \textbf{Total} \\
\midrule
\multirow{2}{*}{English} & Train & 2296 & 2245 & 4541 \\
    & Dev   &  834 &  816 & 1650 \\
\midrule
\multirow{2}{*}{German} & Train & 4924 & 6649 & 11,573 \\
    & Dev   & 1790 & 2418 &  4208 \\
\midrule
\multirow{2}{*}{Spanish} & Train & 5167 & 5383 & 10,550 \\
    & Dev   & 1879 & 1958 &  3837 \\
\midrule
\multirow{2}{*}{Urdu} & Train & 2183 & 2430 &  4613 \\
    & Dev   &  794 &  884 &  1678 \\
\bottomrule
\end{tabular}
\caption{Label distribution in training and development sets across all languages.}
\label{tab:dataset-distribution}
\end{table}

\begin{figure}
    \centering
    \includegraphics[width=1\linewidth]{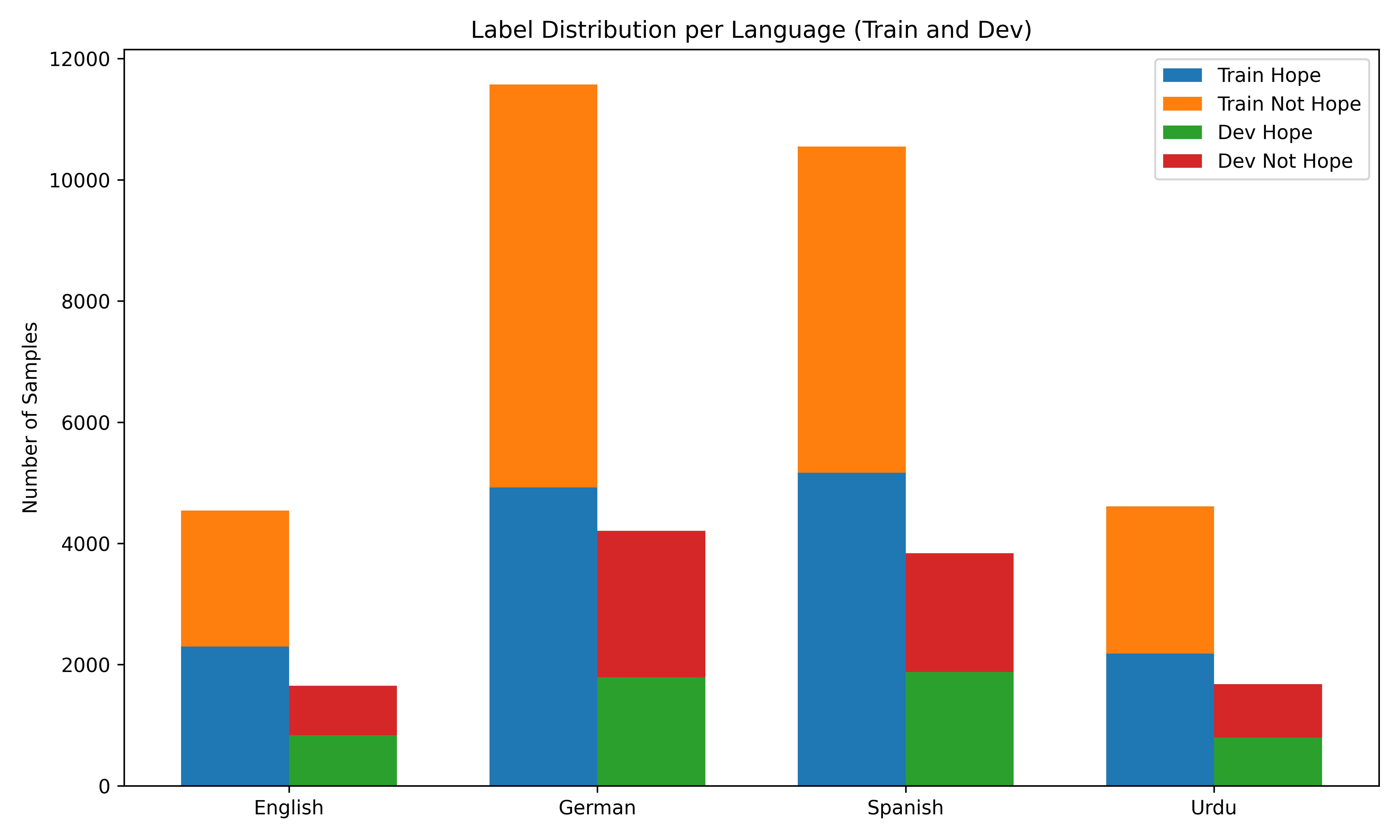}
    \caption{Label Distribution per Language}
    \label{fig:enter-label}
\end{figure}

\begin{figure}
    \centering
    \includegraphics[width=1\linewidth]{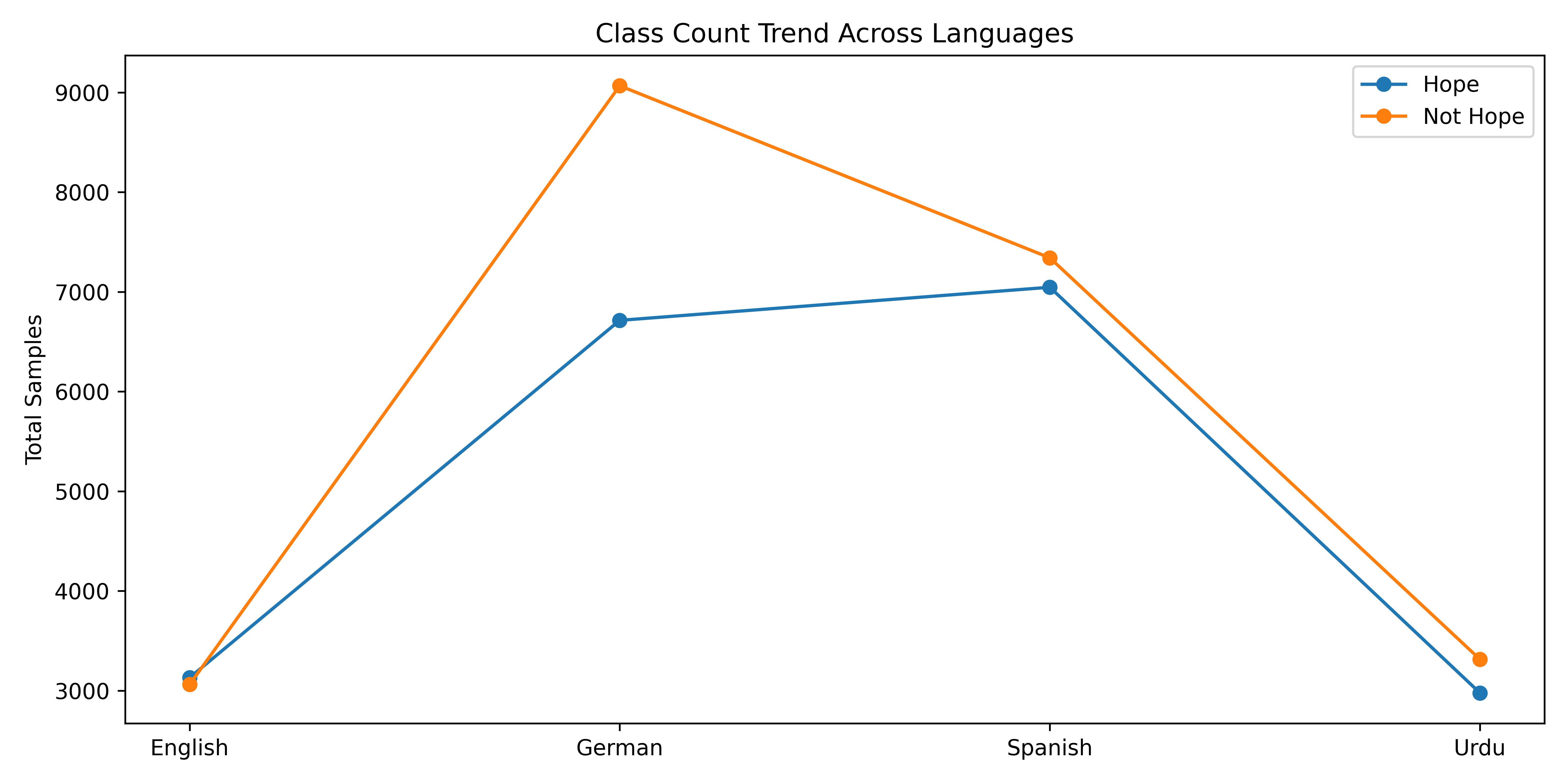}
    \caption{Class Count Trend Across Languages}
    \label{fig:enter-label}
\end{figure}
\noindent
As shown in Figures 1 and 2, all datasets are relatively balanced, though slight variations exist in the ratio of \textit{Hope} to \textit{Not Hope} across languages. German and Spanish corpora are significantly larger compared to English and Urdu. This multilingual setup supports robust evaluation of model performance across high-resource and low-resource languages.

\subsection{Preprocessing}
All languages were in three CSV files tagged train, development and test datasets. We processed all the datasets through text normalisation, including lowercasing, punctuation and URL removal. Additionally, word counts were calculated to assess the average text length per label. The processed texts were tokenised using TF-IDF for classical machine learning models, and pre-trained tokenisers (mBERT and XLM-RoBERTa) for transformer-based models. Categorical labels were mapped to numerical representations for downstream modelling. For consistency, padding and truncation were applied with a maximum sequence length of 128 tokens.

\subsection{Active Learning Strategy}

To enhance the model performance to be effective regardless of the size of the dataset, we implemented an \textbf{Active Learning} framework centred on \textit{uncertainty sampling} on the training dataset as it was split into smaller sizes for the iterative learning. This approach iteratively identifies and selects the most ambiguous samples from the unlabeled pool (in our case, the other split part of the training dataset), thereby increasing the training data as the iteration increases and learning from its error.

\subsubsection{Uncertainty Sampling}
At each iteration, the trained model predicts class probabilities for all instances in the remaining unseen part of the training dataset. We compute the uncertainty of each prediction using the entropy of the output probability distribution:
\[
H(p) = - \sum_{i=1}^{C} p_i \log(p_i)
\]
where \(p_i\) denotes the predicted probability for class \(i\), and \(C\) is the total number of classes (2 in this case). The higher the entropy, the less confident the model is about its prediction.

\subsubsection{Selection and Learning}
In each active learning round, the top 20 samples with the highest entropy scores (i.e., most uncertain predictions) were selected. Since the remaining training dataset the model is being evaluated on is already labelled, we program the model to directly retrieve the ground truth labels. These samples were then appended to the training set. This helps the model keep learning on chunks of the training dataset actively by learning from its errors and increasing accuracy as the iteration increases.

\subsubsection{Iteration and Retraining}
After augmenting the training data with uncertain samples, the model was retrained on the new dataset. This iterative process continued for a fixed number of rounds (typically 3–5) or until performance on the remaining training dataset plateaued. Separate active learning loops were executed for both Logistic Regression and the transformer-based models (mBERT and XLM-RoBERTa), maintaining consistency in selection size and evaluation.

\subsubsection{Benefits}
This strategy allowed the model to focus on learning from difficult examples, thereby improving generalisation while simulating a realistic low-resource scenario. The process was particularly effective for detecting hope speech, which is often subtle and context-dependent, making it ideal for uncertainty-based prioritisation.

\subsection{Model Architecture}
\subsubsection{Logistic Regression}
As a baseline, we trained a traditional machine learning classifier using \textit{Logistic Regression} with \textit{TF-IDF} vectorised features. The input text was transformed into numerical representations using unigram TF-IDF features. The logistic regression model was trained using L2 regularisation and optimised using the liblinear solver. The active learning strategy described previously was applied to iteratively enhance performance by incorporating the most uncertain samples from the development set into the training data.

\subsubsection{Multilingual BERT (mBERT)}
We fine-tuned the \texttt{bert-base-multilingual-cased} model, a transformer architecture pretrained on over 100 languages using masked language modelling. Fine-tuning was performed using the HuggingFace \texttt{Trainer} API. The model was trained for 3 epochs with a learning rate of 2e-5 and a batch size of 32. Tokenisation was handled using the pretrained mbert tokeniser with truncation and padding. A stratified 5-fold cross-validation was used to ensure robustness across different data splits. The previously detailed active learning approach was integrated after each fold, refining the model with iteratively selected uncertain samples.

\subsubsection{XLM-RoBERTa}
We also fine-tuned \texttt{xlm-roberta-base}, a transformer model pretrained on 100 languages using a more diverse CommonCrawl corpus. XLM-RoBERTa employs dynamic masking and is trained on significantly more data than mbert, potentially improving its ability to capture subtle linguistic cues in code-mixed and low-resource contexts. Training followed a similar procedure to mBert, including a learning rate of 2e-5, batch size of 32, and 5-fold cross-validation. Dropout regularisation was employed to reduce overfitting. Active learning, as previously outlined, was applied to incrementally augment the training set with high-uncertainty examples.

\subsection{Evaluation and Prediction}
All models were evaluated on the development dataset and test dataset using classification metrics. Confusion matrices were plotted to visualise performance across classes on the development dataset, as the ground truth label was not released for the test dataset for this task. Final predictions were generated on the unseen test set, and results were saved as CSV files for submission. Both Logistic Regression and transformer-based models provided predictions labelled as “Hope” or “Not Hope”.

\section{Results}
This section presents the evaluation outcomes of the models across different languages, starting with the English dataset. We report performance on both the development and test sets, including classification metrics and confusion matrices to provide detailed insights.

\subsection{Development Set Results}
Table~\ref{tab:dev-results-all} summarizes the classification performance across all datasets (English, German, Spanish, and Urdu) for the development set. Models compared include Logistic Regression (LR), multilingual BERT (mBERT), and XLM-RoBERTa (XLM-R). Overall, the transformer-based models consistently outperform Logistic Regression, with XLM-RoBERTa achieving the highest precision, recall, F1-score, and accuracy in most cases.

\begin{table}[h!]
\centering
\resizebox{\columnwidth}{!}{%
\begin{tabular}{lcccc}
\toprule
\textbf{Model} & \textbf{English} & \textbf{German} & \textbf{Spanish} & \textbf{Urdu} \\
\midrule
Logistic Regression & 0.84 & 0.85 & 0.81 & 0.95 \\
mBERT & 0.87 & 0.87 & 0.83 & 0.98 \\
XLM-RoBERTa & 0.88 & 0.87 & 0.84 & 0.98 \\
\bottomrule
\end{tabular}%
}
\caption{Development set performance across all languages.}
\label{tab:dev-results-all}
\end{table}

The confusion matrix analysis across all languages, as displayed in the appendix B section of this paper shows that transformer models especially XLM-RoBERTa perform better in distinguishing between "Hope" and "Not Hope" classes. Logistic Regression tends to have higher false negatives, particularly for "Hope" instances. mBERT shows a reduction in both false positives and false negatives compared to Logistic Regression. XLM-RoBERTa achieves the best balance, with lower misclassifications and more confident predictions, which explains its superior F1-scores across all languages.

\subsection{Test Set Results}
Table~\ref{tab:test-results-all} presents the performance of all models (Logistic Regression, mBERT, XLM-RoBERTa) on the official test sets for each language. XLM-RoBERTa once again outperforms the other models, particularly in the weighted and macro-averaged F1-scores, across English, German, Spanish, and Urdu datasets.

\begin{table}[h!]
\centering
\resizebox{\columnwidth}{!}{%
\begin{tabular}{lcccc}
\toprule
\textbf{Model} & \textbf{English} & \textbf{German} & \textbf{Spanish} & \textbf{Urdu} \\
\midrule
Logistic Regression & 0.82 & 0.81 & 0.79 & 0.93 \\ 
mBERT & 0.84 & 0.85 & 0.83 & 0.95 \\ 
XLM-RoBERTa & 0.85 & 0.87 & 0.83 & 0.95 \\ 
\bottomrule 
\end{tabular}%
} 
\caption{Test set performance across all languages.} 
\label{tab:test-results-all} 
\end{table}

The results for the test set reveal that, while Logistic Regression continues to show strong recall for the "Not Hope" class, transformer-based models (especially XLM-RoBERTa) demonstrate more accurate identification of "Hope" content with fewer misclassifications across both classes. XLM-RoBERTa excels in both precision and recall, providing the best balance for detecting hopeful content, particularly in the multilingual setting.

Detailed result for each language are shown in tables in the Appendix section of the paper.

\subsection{Cross-Language and Cross-Model Performance Comparison}

Overall, XLM-RoBERTa consistently achieves the highest performance across all languages and metrics, particularly excelling in German and Urdu, where it achieves macro and weighted F1 scores close to or exceeding 0.95. mBERT also performs robustly across all languages, often approaching or slightly trailing behind XLM-RoBERTa. Logistic Regression, while simpler and faster, underperforms compared to transformer-based models, especially in morphologically rich languages such as German and Spanish.

Interestingly, the Urdu dataset demonstrates the highest scores across all models, likely due to clearer language patterns or more consistent annotations. Conversely, Spanish and German appear more challenging, with slightly lower scores, highlighting the complexities of multilingual hope speech detection.

These findings confirm that transformer-based models are more effective in cross-lingual settings, with XLM-RoBERTa being the most capable and generalizable model across diverse languages.
\begin{figure}
    \centering
    \includegraphics[width=1\linewidth]{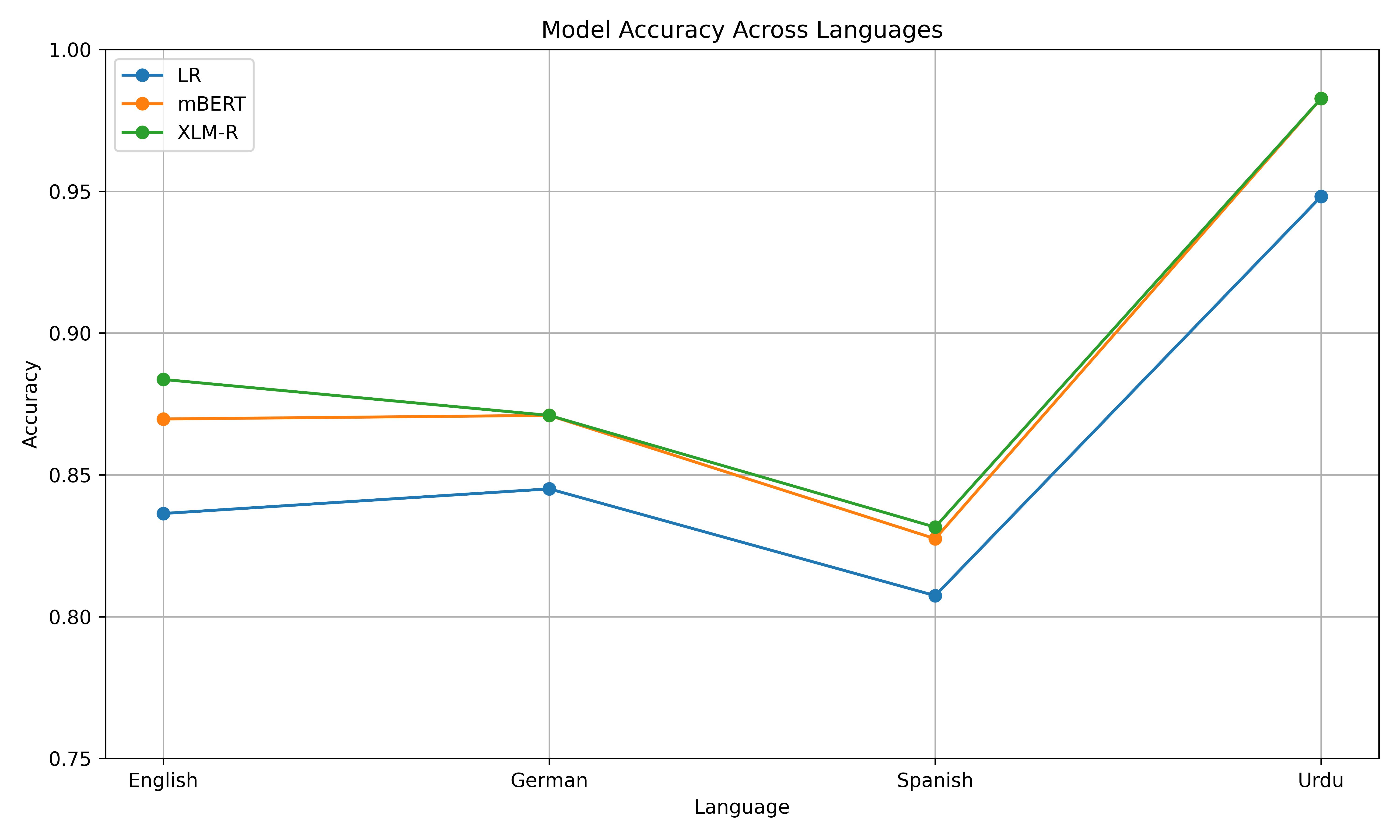}
    \caption{Model Accuracies Across the languages.}
    \label{fig:enter-label}
\end{figure}

\begin{figure}
    \centering
    \includegraphics[width=1\linewidth]{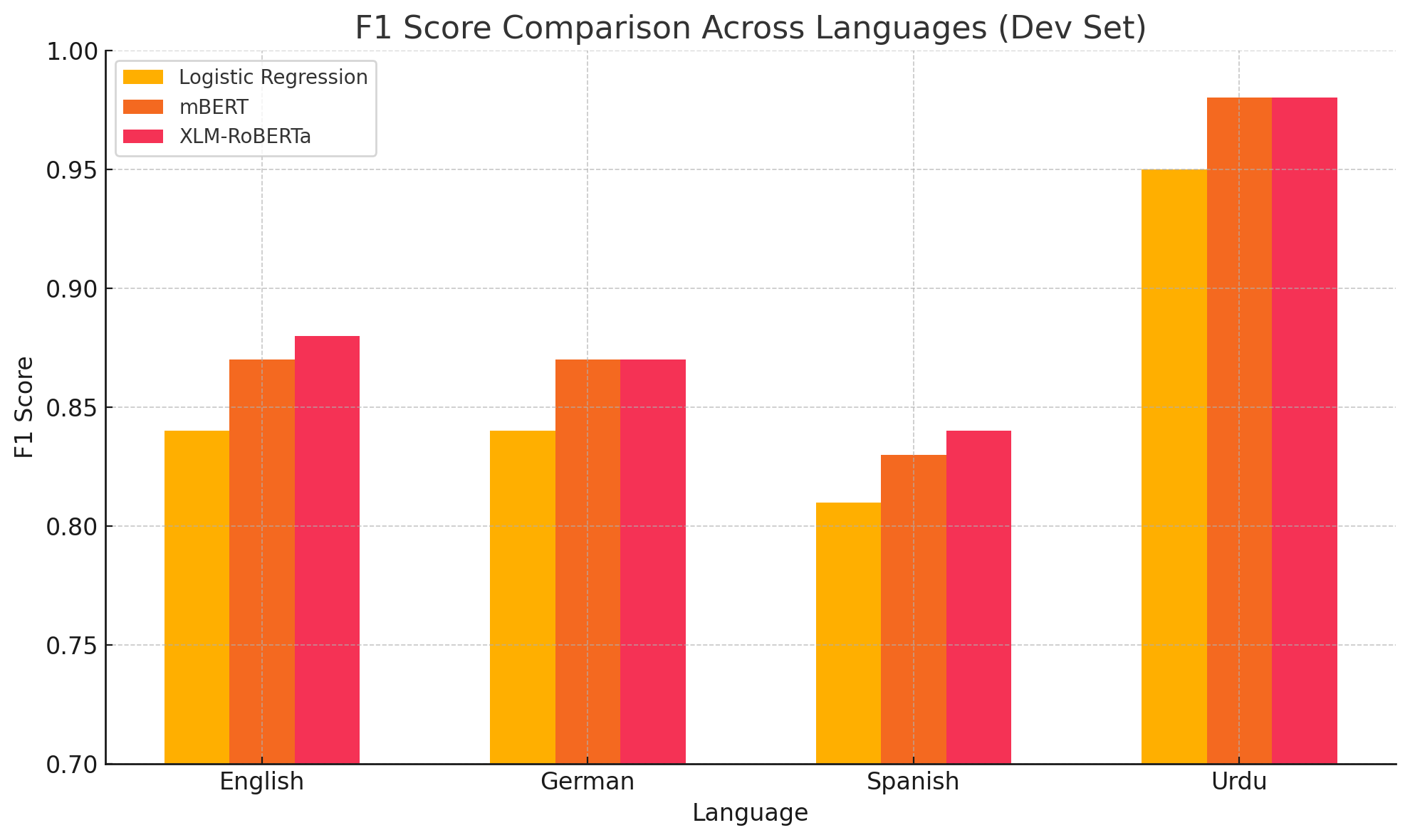}
    \caption{Model F1-score Comparison per Language}
    \label{fig:enter-label}
\end{figure}
Figure 3 and 4 visualises the accuracy and F1-Score of each model across the four languages: English, German, Spanish, and Urdu. By plotting these values, the graph highlights which models perform consistently well across languages and which may be more sensitive to language-specific challenges. It provides a clear, comparative overview of each model’s overall correctness in predictions.

The Appendix B section of this paper contains more graphs with detailed visualisations of the model performances in other metrics.

\subsection{Comparison with Existing Techniques}

Table~\ref{tab:literature-comparison} compares the performance of existing hope speech detection approaches as discussed in the literature with our proposed models. These include a range of traditional machine learning methods, deep learning networks, and transformer-based architectures across multiple languages and datasets. We report the best corresponding results from our models to provide a fair baseline for comparison.

\begin{table*}[ht]
\centering
\small
\renewcommand{\arraystretch}{2}
\begin{tabular}{|p{2.7cm}|p{2.9cm}|p{1.4cm}|p{2.3cm}|p{2.4cm}|p{2.4cm}|}
\hline
\textbf{Reference} & \textbf{Approach} & \textbf{Languages} & \textbf{Dataset} & \textbf{Reported Result} & \textbf{Ours (Best)} \\
\hline
\cite{sharma2025ensemble} & Ensemble (LSTM + mBERT + XLM-R) & EN, KN, ML, TA & HOPE 2023 & EN: 0.93, KN: 0.74, ML: 0.82, TA: 0.60 & EN: 0.84 \\
\hline
\cite{divakaran2024hope} & TF-IDF, mBERT, ML + TL models & EN, ES & HOPE@IberLEF 2024 & ES (Task 2.a): 0.82 (Macro F1) & ES: 0.84 \\
\hline
\cite{ahmad2025multilingual} & BERT-based with translation & EN, UR & In-house corpus & EN: 87\% Acc, UR: 79\% Acc & EN: 0.84, UR: 0.98 \\
\hline
\cite{eyob2024enhancing} & BERT, SVM, RF & EN & HOPE@IberLEF 2024 & EN: 0.85 (Macro F1) & EN: 0.84 \\
\hline
\textbf{Ours (XLM-R)} & XLM-RoBERTa fine-tuned & EN, ES, DE, UR & HOPE Dev + Test Sets & \textbf{F1: 0.84–0.98} & --- \\
\hline
\textbf{Ours (mBERT)} & mBERT fine-tuned & EN, ES, DE, UR & HOPE Dev + Test Sets & F1: 0.83–0.98 & --- \\
\hline
\textbf{Ours (LR)} & Logistic Regression (TF-IDF) & EN, ES, DE, UR & HOPE Dev + Test Sets & F1: 0.80–0.95 & --- \\
\hline
\end{tabular}
\caption{Comparison of our models with previously published approaches. EN: English, ES: Spanish, DE: German, UR: Urdu, KN: Kannada, ML: Malayalam, TA: Tamil.}
\label{tab:literature-comparison}
\end{table*}

\noindent The table illustrates that our multilingual transformer models particularly XLM-RoBERTa achieve comparable or superior performance across languages. While some ensemble models perform well on specific high-resource languages like English, our models generalize effectively even in low-resource settings such as Urdu and Spanish. Notably, our XLM-R model achieves a weighted F1-score of up to 0.98 on Urdu and 0.84 on Spanish, outperforming or matching prior state-of-the-art results.

\section{Conclusion}

This study investigated multilingual hope speech detection using both classical and transformer-based models, Logistic Regression, mBERT, and XLM-RoBERTa across English, Spanish, German, and Urdu. A key innovation was the integration of active learning to maintain strong model performance with small annotated data. Results from the HOPE shared task datasets showed that transformer models, especially XLM-RoBERTa, outperformed traditional approaches, with particularly high effectiveness in low-resource languages like Urdu and Spanish. Compared to existing methods, our approach demonstrated competitive or superior results even with less labelled data, highlighting the value of combining transfer and active learning for scalable, real-world hope speech detection.
\section{Limitations}
Despite promising results, this study has several limitations. The language coverage is limited to English, Spanish, German, and Urdu, excluding other low-resource languages like Swahili or Amharic, and the datasets may not represent full dialectal or cultural diversity. Additionally, the binary classification (hope vs. non-hope) oversimplifies the complexity of hopeful expressions, which could be better captured through multi-class or regression approaches. The models were trained on social media data, which limits their generalizability to other domains. Transformer-based models like mBERT and XLM-RoBERTa, while effective, are resource-intensive, making real-time or low-resource deployment challenging. Furthermore, their black-box nature limits interpretability, raising ethical concerns about misclassification impacting content visibility or user profiling. Future work should address these issues to develop more robust, inclusive, and ethically sound systems for hope speech detection.

\section*{Acknowledgments}
The work was done with partial support from the Mexican Government through the grant A1-S-47854 of CONACYT, Mexico,
grants 20241816, 20241819, and 20240951 of the Secretaría de Investigación y Posgrado of the Instituto Politécnico Nacional, Mexico. The authors thank the CONACYT for the computing
resources brought to them through the Plataforma de Aprendizaje Profundo para Tecnologías del Lenguaje of the Laboratorio de Supercómputo of the INAOE, Mexico and acknowledge the support
of Microsoft through the Microsoft Latin America PhD Award.

\bibliographystyle{acl_natbib}
\bibliography{ranlp2023}

\appendix
\section{Appendix}
\begin{table}[hbtp]
\centering
\resizebox{\linewidth}{!}{
\begin{tabular}{lcccc}
\toprule
\textbf{Model} & \textbf{Precision} & \textbf{Recall} & \textbf{F1-score} & \textbf{Accuracy} \\
\midrule
Logistic Regression & 0.84 & 0.84 & 0.84 & 0.84 \\
mBERT               & 0.87 & 0.87 & 0.87 & 0.87 \\
XLM-RoBERTa         & 0.88 & 0.88 & 0.88 & 0.88 \\
\bottomrule
\end{tabular}%
}
\caption{English development set performance.}
\label{tab:dev-eng}
\end{table}

\begin{table}[hbtp]
\centering
\resizebox{\linewidth}{!}{
\begin{tabular}{lcccc}
\toprule
\textbf{Model} & \textbf{Precision} & \textbf{Recall} & \textbf{F1-score} & \textbf{Accuracy} \\
\midrule
Logistic Regression & 0.84 & 0.84 & 0.84 & 0.85 \\
mBERT               & 0.87 & 0.87 & 0.87 & 0.87 \\
XLM-RoBERTa         & 0.88 & 0.88 & 0.87 & 0.87 \\
\bottomrule
\end{tabular}%
}
\caption{German development set performance.}
\label{tab:dev-ger}
\end{table}

\begin{table}[h!]
\centering
\resizebox{\linewidth}{!}{
\begin{tabular}{lcccc}
\toprule
\textbf{Model} & \textbf{Precision} & \textbf{Recall} & \textbf{F1-score} & \textbf{Accuracy} \\
\midrule
Logistic Regression & 0.81 & 0.81 & 0.81 & 0.81 \\
mBERT               & 0.83 & 0.83 & 0.83 & 0.83 \\
XLM-RoBERTa         & 0.84 & 0.84 & 0.84 & 0.84 \\
\bottomrule
\end{tabular}%
}
\caption{Spanish development set performance.}
\label{tab:dev-spanish}
\end{table}

\begin{table}[h!]
\centering
\resizebox{\linewidth}{!}{
\begin{tabular}{lcccc}
\toprule
\textbf{Model} & \textbf{Precision} & \textbf{Recall} & \textbf{F1-score} & \textbf{Accuracy} \\
\midrule
Logistic Regression & 0.95 & 0.95 & 0.95 & 0.95 \\
mBERT               & 0.98 & 0.98 & 0.98 & 0.98 \\
XLM-RoBERTa         & 0.98 & 0.98 & 0.98 & 0.98 \\
\bottomrule
\end{tabular}%
}
\caption{Urdu development set performance.}
\label{tab:dev-urdu}
\end{table}

\begin{table}[h!]
\centering
\resizebox{\linewidth}{!}{
\begin{tabular}{lcccc}
\toprule
\textbf{Model} & \textbf{Weighted F1} & \textbf{Macro F1} & \textbf{Accuracy} \\
\midrule
Logistic Regression & 0.821 & 0.821 & 0.821 \\
mBERT               & 0.841 & 0.841 & 0.841 \\
XLM-RoBERTa         & 0.851 & 0.851 & 0.851 \\
\bottomrule
\end{tabular}%
}
\caption{English test set performance.}
\label{tab:test-eng}
\end{table}

\begin{table}[h!]
\centering
\resizebox{\linewidth}{!}{
\begin{tabular}{lcccc}
\toprule
\textbf{Model} & \textbf{Weighted F1} & \textbf{Macro F1} & \textbf{Accuracy} \\
\midrule
Logistic Regression & 0.811 & 0.811 & 0.811 \\
mBERT               & 0.851 & 0.850 & 0.849 \\
XLM-RoBERTa         & 0.867 & 0.864 & 0.867 \\
\bottomrule
\end{tabular}%
}
\caption{German test set performance.}
\label{tab:test-ger}
\end{table}

\begin{table}[h!]
\centering
\resizebox{\linewidth}{!}{
\begin{tabular}{lcccc}
\toprule
\textbf{Model} & \textbf{Weighted F1} & \textbf{Macro F1} & \textbf{Accuracy} \\
\midrule
Logistic Regression & 0.801 & 0.801 & 0.791 \\
mBERT               & 0.821 & 0.830 & 0.829 \\
XLM-RoBERTa         & 0.832 & 0.832 & 0.832 \\
\bottomrule
\end{tabular}%
}
\caption{Spanish test set performance.}
\label{tab:test-spanish}
\end{table}

\begin{table}[h!]
\centering
\resizebox{\linewidth}{!}{
\begin{tabular}{lcccc}
\toprule
\textbf{Model} &  \textbf{Weighted F1} & \textbf{Macro F1} & \textbf{Accuracy} \\
\midrule
Logistic Regression & 0.930 & 0.930 & 0.930 \\
mBERT               & 0.945 & 0.946 & 0.947 \\
XLM-RoBERTa         & 0.950 & 0.949 & 0.950 \\
\bottomrule
\end{tabular}%
}
\caption{Urdu test set performance.}
\label{tab:test-urdu}
\end{table}

\onecolumn
\section{Appendix}
\begin{figure}[ht]
\centering
\includegraphics[width=1\textwidth]{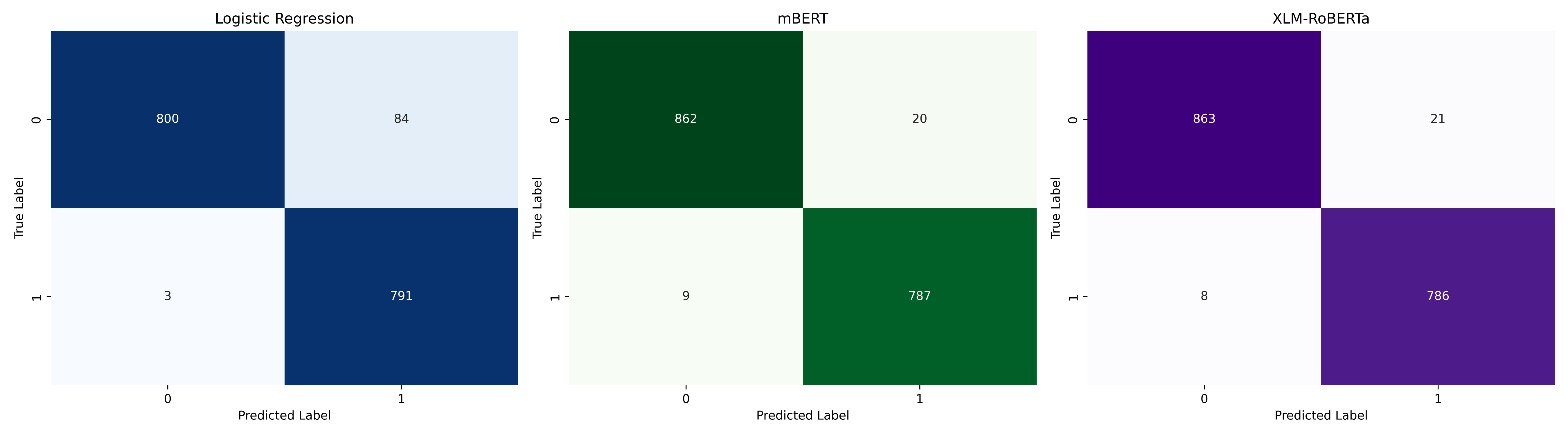}
\caption{Combined confusion matrix for Logistic Regression, mBERT, and XLM-RoBERTa models on the Urdu development set. The matrices are placed side by side for comparison.}
\label{fig:combined_confusion_matrices_urdu_dev}
\end{figure}
\begin{figure}[ht]
\centering
\includegraphics[width=1\textwidth]{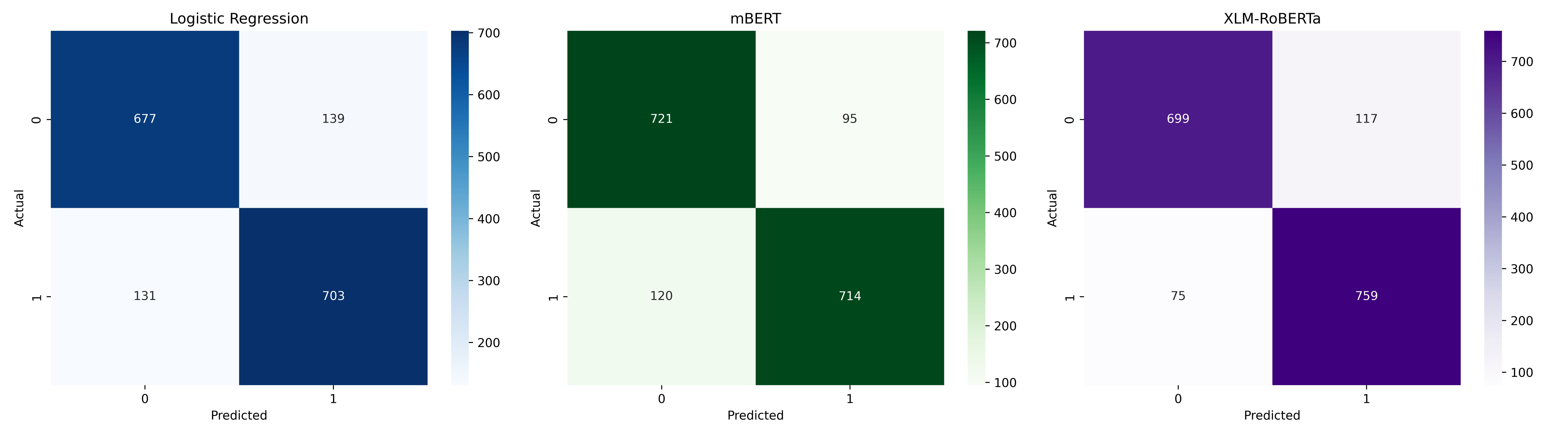}
\caption{Combined confusion matrix for Logistic Regression, mBERT, and XLM-RoBERTa models on the English development set. The matrices are placed side by side for comparison.}
\label{fig:combined_confusion_matrices_eng_dev}
\end{figure}
\begin{figure}[ht]
\centering
\includegraphics[width=1\textwidth]{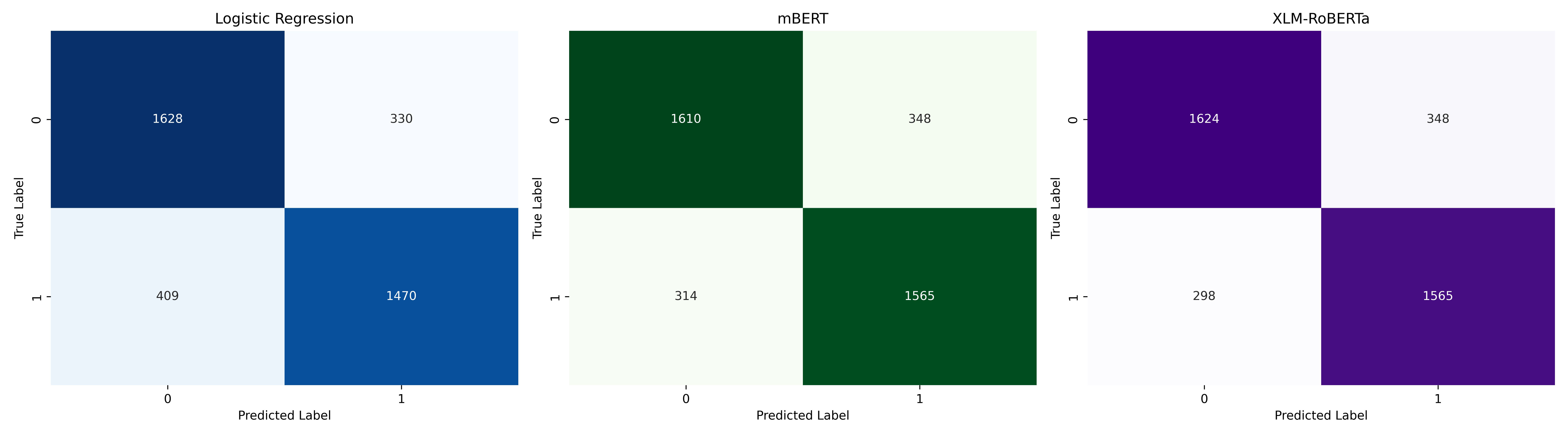}
\caption{Combined confusion matrix for Logistic Regression, mBERT, and XLM-RoBERTa models on the Spanish development set. The matrices are placed side by side for comparison.}
\label{fig:combined_confusion_matrices_esp_dev}
\end{figure}
\begin{figure}[ht]
\centering
\includegraphics[width=1\textwidth]{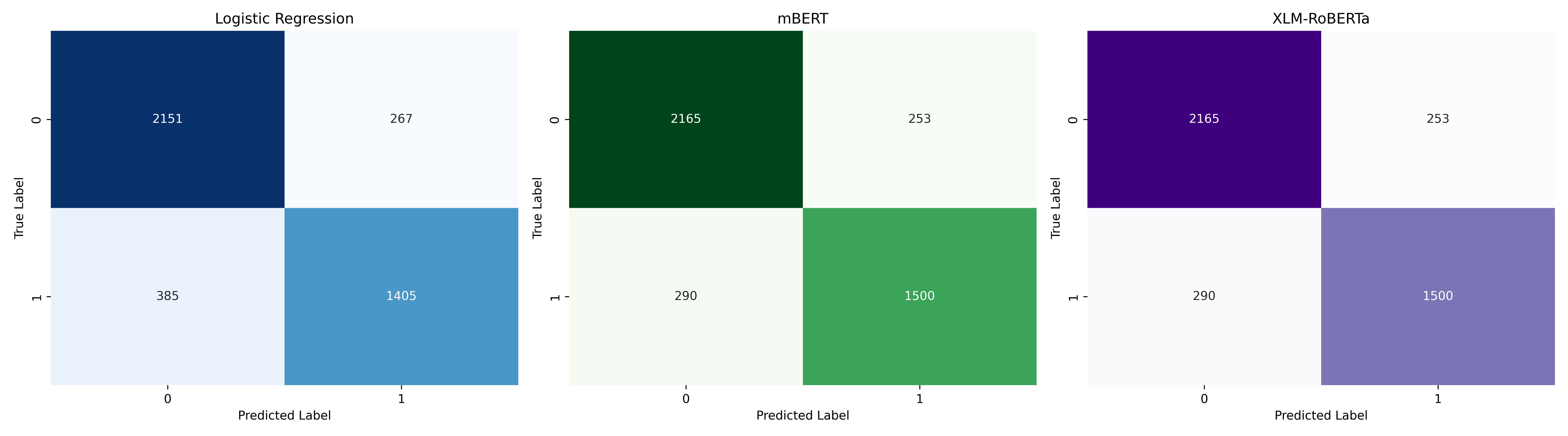}
\caption{Combined confusion matrix for Logistic Regression, mBERT, and XLM-RoBERTa models on the German development set. The matrices are placed side by side for comparison.}
\label{fig:combined_confusion_matrices_ger_dev}
\end{figure}[ht]
\begin{figure}
    \centering
    \includegraphics[width=1\linewidth]{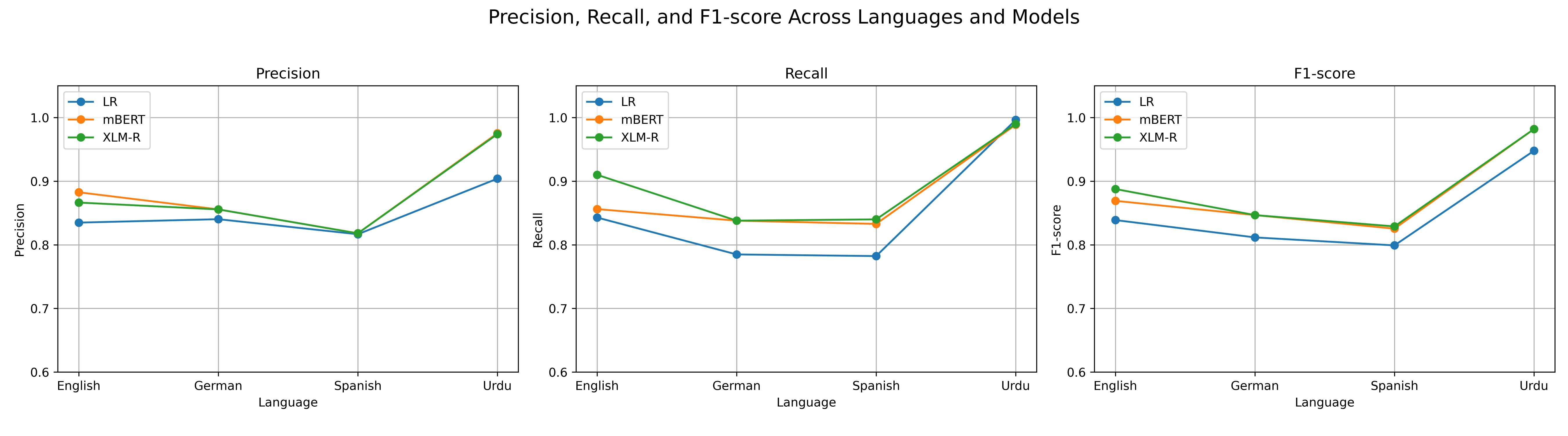}
    \caption{Precision, Recall, and F1-score for each model across the four languages}
    \label{fig:enter-label}
\end{figure}

\end{document}